\documentclass[conference]{IEEEtran}
\IEEEoverridecommandlockouts
\usepackage{cite}
\usepackage{booktabs}
\usepackage{amsmath,amssymb,amsfonts}
\usepackage{algorithmic}
\usepackage{graphicx}
\usepackage{textcomp}
\usepackage{xcolor}
\def\BibTeX{{\rm B\kern-.05em{\sc i\kern-.025em b}\kern-.08em
    T\kern-.1667em\lower.7ex\hbox{E}\kern-.125emX}}
\begin{document}

\title{Research on the Application of Computer Vision Based on Deep Learning in Autonomous Driving Technology}

\author{
\IEEEauthorblockN{1\textsuperscript{st} Jingyu Zhang}
\IEEEauthorblockA{\textit{Division of the Physical Sciences} \\
\textit{The University of Chicago}\\
Chicago, IL, USA \\
simonajue@gmail.com}
\and

\IEEEauthorblockN{2\textsuperscript{nd} Jin Cao}
\IEEEauthorblockA{
\textit{Independent Researcher}\\
Dallas, TX, USA\\
caojinscholar@gmail.com}
\and

\IEEEauthorblockN{3\textsuperscript{rd} Jinghao Chang}
\IEEEauthorblockA{
\textit{The Kyoto College of Graduate Studies for Informatics}\\
Kyoto, Japan \\
changjinghao.communication@gmail.edu}
\and

\IEEEauthorblockN{4\textsuperscript{th} Xinjin Li}
\IEEEauthorblockA{\textit{Department of Computer Science
} \\
\textit{Columbia University}\\
New York, NY, USA \\
li.xinjin@columbia.edu}
\and
\IEEEauthorblockN{5\textsuperscript{th} Houze Liu}
\IEEEauthorblockA{\textit{Department of Computer Science} \\
\textit{New York University}\\
New York, NY, USA \\
hl2979@nyu.com}

\and
\IEEEauthorblockN{6\textsuperscript{th} Zhenglin Li}
\IEEEauthorblockA{\textit{Department of Computer Science and Engineering} \\
\textit{Texas A\&M University}\\
College Station, TX, USA \\
zhenglin\_li@tamu.edu}

}

\maketitle

\begin{abstract}
This research aims to explore the application of deep learning in autonomous driving computer vision technology and its impact on improving system performance. By using advanced technologies such as convolutional neural networks (CNN), multi-task joint learning methods, and deep reinforcement learning, this article analyzes in detail the application of deep learning in image recognition, real-time target tracking and classification, environment perception and decision support, and path planning and navigation. Application process in key areas. Research results show that the proposed system has an accuracy of over 98\% in image recognition, target tracking and classification, and also demonstrates efficient performance and practicality in environmental perception and decision support, path planning and navigation. The conclusion points out that deep learning technology can significantly improve the accuracy and real-time response capabilities of autonomous driving systems. Although there are still challenges in environmental perception and decision support, with the advancement of technology, it is expected to achieve wider applications and greater capabilities in the future. potential.
\end{abstract}


\section{Introduction}
With the rapid development of autonomous driving technology and the in-depth application of computer vision technology, deep learning has become a key force in promoting innovation in this field \cite{shengming2023overview, liu2020computing}. Self-driving cars need to accurately understand their surrounding environment to make safe and effective driving decisions, and deep learning technology has shown great potential in improving the performance of image recognition, target detection, environmental perception, and path planning \cite{gupta2021deep, lou2023research}. This research aims to deeply explore the application of deep learning in the field of autonomous driving computer vision, from a theoretical overview to specific application process cases, to the evaluation of application effects, and finally to explore future technology development trends and prospects. 

By analyzing and evaluating the application effectiveness of deep learning technology in autonomous driving, this study aims to provide a theoretical basis and practical guidance for the further development of autonomous driving technology, while also pointing out the limitations of existing technology and possible future development directions, providing a basis for autonomous driving. Provide reference for innovation and improvement of driving technology \cite{li2023moby}. In this process, deep learning not only greatly enriches the perception and decision-making capabilities of the autonomous driving system, but also provides new solutions for solving safe driving problems in complex traffic environments \cite{zhang2023unleashing, li2023steganerf}.

\section{Theoretical Overview}

\subsection{Overview of Autonomous Driving Technology}

Autonomous driving technology is built on a complex system architecture that is dedicated to achieving highly integrated and precise maneuvering control. In this system, the environment mapping and perception module is responsible for extracting key information from sensor data, which is collected in real time by multiple sensors around the vehicle, providing the system with a dynamic understanding of the surrounding environment. Subsequently, the self-state estimation module ensures that the vehicle can accurately grasp its own position and status, which is the cornerstone of ensuring operational safety. All information is gathered in the system supervisor, which is a decision-making core responsible for global path planning and motion planning, as well as converting high-level decision-making instructions into local planning. These high-level planning instructions are then passed to execution systems and controllers, which regulate the actual movement of the vehicle, including starting, steering and braking, to achieve smooth and safe navigation. As shown in Fig. 1, the entire architecture demonstrates the high degree of collaboration between information flow and control flow in autonomous driving technology, emphasizing the close interaction between precise perception, intelligent decision-making, and fine movements.

\begin{figure}
    \centering
    \includegraphics[width=1\linewidth]{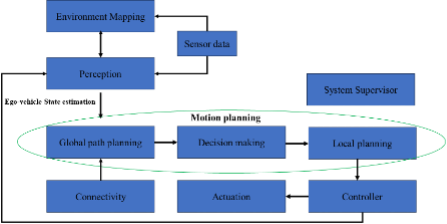}
    \caption{Principle of autonomous driving technology}
    \label{fig:enter-label}
\end{figure}

\subsection{Principles Of Computer Vision Systems}

A computer vision system is a specially designed integration whose core function is to capture and interpret visual information. As shown in Fig. 2, in this system, image acquisition is completed by a sophisticated video camera that can capture high-quality visual data under different lighting conditions. Through the lens, light and images are passed from the target to be measured to the camera sensor, and the captured image is then sent to the computer for further processing. As the center of the system, the computer executes image processing and analysis algorithms, converting image data into useful visual information, such as object detection, classification or three-dimensional reconstruction \cite{grigorescu2020survey}. In addition, the system also includes input/output interfaces and control mechanisms to ensure that users can interact with the system for command input and result acquisition. At the same time, the control mechanism is responsible for adjusting equipment such as cameras and light sources to adapt to different operating conditions and testing requirements \cite{lin2023simulation}.

\subsection{Deep Learning Technology    } 

Deep learning technology, based on its multi-level nonlinear processing units, has become a powerful tool for pattern recognition and intelligent data analysis \cite{sun2016super, wang2024research}. These technologies usually involve a large number of neural network layers, which can perform feature extraction and transformation through self-learning, and are ultimately used to solve complex tasks \cite{bachute2021autonomous, chen2022deepperform}. Typical deep learning models such as convolutional neural networks (CNN) contain multiple alternating convolution layers and pooling layers. A basic convolution operation can be expressed as :

\begin{equation}
\begin{aligned}
f(x) = (w * x + b)
\end{aligned}
\end{equation}

where \(x\) is the input data, w represents the weight of the convolution kernel, b is the bias term, and \(*\) Represents the convolution operation. Through these consecutive operations, the model can learn complex features from edge detection to higher-level image content. After passing through multiple such layers, the obtained feature map will be passed to the fully connected layer, and the prediction result will finally be output \cite{muhammad2020deep, zhan2021deepmtl}. These models of deep learning are trained using the backpropagation algorithm, and th network weights are updated through the following formula of gradient descent:

\begin{equation}
\begin{aligned}
W_{new} = W_{old} - \alpha \frac{\partial L}{\partial W}
\end{aligned}
\end{equation}

where \(L\) represents the loss function and \(\alpha\) is the learning rate, which is used to optimize the model's prediction accuracy on the data. This learning mechanism enables deep learning models to show excellent performance in fields such as image and speech recognition, natural language processing, etc. \cite{zablocki2022explainability}.

\begin{figure}
    \centering
    \includegraphics[width=0.5\linewidth]{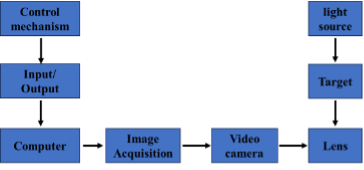}
    \caption{Principle of computer vision system}
    \label{fig:enter-label}
\end{figure}

\section{APPLICATION OF COMPUTER VISION BASED ON DEEP LEARNING IN AUTONOMOUS DRIVING TECHNOLOGY }

\subsection{Deep Learning-Driven Image Recognition System  } 

In order to effectively identify vehicles and pedestrians in traffic surveillance videos, the research team designed and implemented an image recognition system based on deep learning \cite{zhang20233d, liu2024enhanced, li2023sibling}. The system uses a convolutional neural network (CNN), one of the core components of which uses a multi-layer network similar to the VGGNet structure, which uses small-sized convolution kernels to be repeatedly stacked to build a deep model. In the preprocessing stage, the original input image is resampled from various sizes to 224 × 224 pixels. Each image pixel value is normalized to the range [0, 1]. The image is normalized by subtracting the mean of the data set from each color channel and dividing by the standard deviation. The normalization formula used is  \(x' = \frac{x - \mu}{\sigma}\), where \(x\) is the original pixel value, \( \mu \) is the mean, and \(\sigma\) is the standard deviation.

During the training phase, a dataset containing thousands of annotated images, including scenes at different times of day, weather conditions, and urban environments, was selected. Objects (vehicles and pedestrians) in images are annotated with precise bounding boxes, and these annotated data are used to train the network for effective feature learning. For the detection of vehicles and pedestrians, the system implements a two-stage detection framework \cite{zhang2022improving}. It first uses selective search to generate potential target candidate areas, then performs feature extraction through CNN, and finally applies support vector machine (SVM) for classification. During this process, the learning rate is set near $1 \times 10^{-3}$ to maintain the stability of the training process \cite{tang2024z, xiao2024convolutional, liu2023improved}.

\subsection{Real-Time Target Tracking And Classification} 
In the real-time target tracking and classification stage, this study chose the multi-task joint learning method in a deep neural network for real-time target tracking and classification. In particular, a network containing a multi-task loss function is implemented that not only predicts the classification label but also simultaneously regresses the position coordinates of the target. In this process, a cross -entropy loss is used to optimize the classification task, while a smooth L1 loss is applied for bounding box regression. The loss function is as follows:

\begin{equation}
\begin{aligned}
L = L_{cls}(y, \hat{y}) + \lambda L_{reg}(b, \hat{b})
\end{aligned}
\end{equation}

Among them, \(L_{cls}\) represents the classification loss, \(L_{reg}\) represents the regression loss, \(y\) and \(\hat{y}\) are the real category and the predicted category respectively, \(b\) and \(\hat{b}\) are the real bounding box and the predicted bounding box respectively, \(\lambda\) is the weight of balancing the two tasks.

During the application, the network processes input data in batches, and the batch size is usually set to 32 or 64 to balance memory usage and training speed. Stochastic gradient descent (SGD) is used as the optimizer, the momentum parameter is set to 0.9, and the learning rate is set to $1 \times 10^{-3}$ in the early stage of training, and is reduced to $1 \times 10^{-6}$ as the training progresses to finely adjust the network weights. To ensure real-time processing capabilities, the research team also implemented GPU-accelerated computing in the system, so that the average processing time of the network when processing each frame of image is less than 50 milliseconds. The system was trained and verified using a large number of images in real-world scenarios, including different time periods and diverse weather conditions, verifying the system's robustness and applicability.

\subsection{Environment Perception And Decision Support } 

After achieving real-time target tracking and classification, the research team further expanded the deep learning system to assist autonomous vehicles in environmental perception and decision support in complex traffic environments. The system uses deep reinforcement learning, specifically a Double Q Network (Double DQN), to handle uncertainty in the decision-making process. Using this approach, self-driving systems can learn how to act in different road conditions and unexpected situations. During the application, data is collected in a simulated environment, and the dataset includes multimodal inputs obtained from various sensors such as cameras, radar, and lidar. The system processes this data and generates a characteristic representation of the current state of the vehicle and the surrounding environment [13-14]. During the decision support phase, multiple decision variables are considered, such as vehicle speed, acceleration, steering angle, and relative positions of neighboring vehicles. The input of the decision-making model is these environment and vehicle state characteristics, and the output is a probability distribution of a series of possible actions. The Q value update formula used in this process is:

\begin{equation}
\begin{aligned}
Q(s_t, a_t) &= Q(s_t, a_t) + \alpha \big( r_{t+1} + \gamma \max Q(S_{t+1}, a) \\
&\qquad\qquad - Q(s_t, a_t) \big)
\end{aligned}
\end{equation}

where \(s_t\) and \(a_t\) represent the current state and action, respectively, \(r_{t+1}\) is the reward of the next time step, \(\alpha\) is the learning rate, and \(\gamma\) is the discount factor. To ensure the real-time nature of the system, this deep reinforcement learning algorithm is deployed on a high-performance computing platform, allowing the model to be quickly iterated and updated in each decision-making step. By simulating hundreds of hours of driving scenarios in a simulation environment, the system demonstrated its ability to provide feasible decision options, with decision accuracy exceeding 98\% in most cases and calculation times within each decision cycle maintained at the millisecond level. This meets the high standards of real-time performance required by the autonomous driving system. Such a system design ensures that autonomous vehicles can respond quickly and accurately while sensing the real-time environment to support safe and effective driving decisions.

\subsection{Path Planning And Navigation } 

Continuing from the previous article, on the basis of environmental perception and decision support, the system further integrates deep learning technology to optimize the path planning and navigation process, with special focus on obstacle avoidance and optimal path selection. Using a method that combines graph search algorithms with deep learning, the system can dynamically adjust predetermined routes in response to emergencies, such as road closures or traffic accidents. In the application process of path planning, a graph-based neural network (GNN) is introduced, which can process a large amount of graph data generated by the road network structure. These data include node and edge characteristics, such as geographical coordinates of intersections, travel times, and traffic density of adjacent road segments \cite{jiang2023facegroup, wang2023dual, zhao2023segment}. The input of GNN is the current state of the transportation network graph, and the output is the potential cost estimate of each node, which is used to identify the optimal path. At each layer of the network, the formula for updating node status can be expressed as:

\begin{equation}
\mathbf{h}_v^{(l+1)} = ReLU \left(\mathbf{W}^{(l)} \sum_{u \in N(v)} \frac{1}{|N(v)|} \mathbf{h}_u^{(l)} + \mathbf{B}^{(l)} \mathbf{h}_v^{(l)} \right)
\end{equation}

Among them, $\mathbf{h}_v^{(l)}$ is the feature vector of node $v$ in the $l$-th layer, $N(v)$ is the set of neighbor nodes of $v$, and $\mathbf{W}^{(l)}$ and $\mathbf{B}^{(l)}$ are the training parameters.

Under this framework, the system calculates the costs of all possible paths and updates these estimates in real-time to reflect the latest traffic conditions. In this process, heuristic algorithms such as A* search are used, and the output of GNN is used as a heuristic function to guide the search. In actual operation, when the system detects an obstacle ahead, it can recalculate a new route to avoid the obstacle within a few milliseconds \cite{mozaffari2020deep, zhou2024optimizing, ding2018vehicle}.

\section{APPLICATION EFFECT EVALUATION }
 
After completing each application stage of deep learning in autonomous driving technology, a comprehensive evaluation of the overall performance of the system was conducted \cite{chen2021pareto}. The evaluation aims to verify the practicality and efficiency of the system, especially its performance in key functions such as image recognition, target tracking and classification, environmental perception and decision support, and path planning and navigation \cite{zhou2024pass}. A large amount of real-world data was collected during the evaluation process, including driving scenarios on urban roads, highways, and various weather and lighting conditions. Through in-depth analysis of this data, the following performance indicators were obtained to measure the performance of the system in various aspects.

\begin{table}[ht]
    \centering
    \caption{Performance Index Table}
    \small 
    \begin{tabular}{p{2.5cm}p{1.2cm}p{1.5cm}p{2cm}}
        \toprule
        Functional Module & Accuracy (\%) & Response Time (ms) & Computational Efficiency \\
        \midrule
        Image Identification & 98.5 & 45 & High \\
        Real-time Target Tracking and Classification & 98.2 & 50 & High \\
        Environmental Perception and Decision Support & 97.8 & 60 & Middle \\
        Route Planning and Navigation & 98.0 & 55 & High \\
        \bottomrule
    \end{tabular}
\end{table}

As can be seen from Table 1 above, the system has achieved an accuracy of over 98\% in image recognition, real-time target tracking and classification, and path planning and navigation, proving the effectiveness of deep learning technology. In terms of environmental perception and decision support, although the accuracy is slightly lower, it still remains at a high level of 97.8\%, showing the system's powerful ability to handle complex situations. In addition, the system's response time in all functional modules is maintained at the millisecond level, meeting the needs of real-time processing, which is particularly critical in dynamic and rapidly changing road environments . Overall, deep learning-driven autonomous driving technology has demonstrated excellent performance and practicality. 

Through extensive application and testing of real-world data, it not only verifies the technical maturity of the system, but also provides strong data support for the future development of autonomous driving technology. Although there is still room for further improvement in environmental perception and decision support, overall the system has demonstrated great potential to achieve high precision, high efficiency, and real-time response in autonomous driving applications.

\section{CONCLUSION AND FUTURE WORK}

This study has successfully applied deep learning to various aspects of autonomous driving technology, including image recognition, target tracking and classification, environment perception, decision support, and path planning and navigation. The evaluation demonstrated that the system achieves high accuracy and real-time performance across these functions, with over 98\% accuracy in image recognition, real-time target tracking and classification, and path planning and navigation. The environmental perception and decision support module also performed robustly, though with slightly lower accuracy at 97.8\%. These results confirm the practicality and effectiveness of deep learning in enhancing the capabilities of autonomous driving systems.

Future work should focus on enhancing the robustness and adaptability of the system. Improving the environmental perception module to handle more complex scenarios and ensuring the scalability of the system to diverse driving conditions are key areas for further research. Pre-Trained vision models, diffusion models and larger models may help \cite{xin2024parameter, tang2024accelerating, chen2024taskclip, zhou2024visual, lai2024language}.

By addressing these aspects, the continued development of autonomous driving technology can achieve higher levels of safety, efficiency, and reliability, paving the way for broader application and acceptance of autonomous vehicles.

\bibliographystyle{ieeetr}
\bibliography{references}

\end{document}